# Semantic Social Network Analysis

Guillaume ERETEO
Orange Labs
Sophia Antipolis, 06921, France
+334 92 94 53 84
guillaume.ereteo@orange-ftgroup.com

Fabien GANDON
Olivier CORBY
EDELWEISS, INRIA
Sophia-Antipolis, France
+334 92 38 77 88
firstname.name@sophia.inria.fr

Michel BUFFA
KEWI, I3S,
Université of Nice, France
+334 92 04 66 60
buffa@unice.fr

**Abstract:** Social Network Analysis (SNA) tries to understand and exploit the key features of social networks in order to manage their life cycle and predict their evolution. Increasingly popular web 2.0 sites are forming huge social network. Classical methods from social network analysis (SNA) have been applied to such online networks. In this paper, we propose leveraging semantic web technologies to merge and exploit the best features of each domain. We present how to facilitate and enhance the analysis of online social networks, exploiting the power of semantic social network analysis.

## 1. INTRODUCTION

Since its birth, the web provided many ways of interacting between us [6], revealing huge social network structures [17], a phenomenon amplified by web 2.0 applications [11]. Researchers extracted social networks from emails, mailing-list archives, hyperlink structure of homepages, co-occurrence of names in documents and from the digital traces created by web 2.0 application usages [9]. Facebook, LinkedIn or Myspace provide huge amounts of structured network data. The emergence of the semantic web approaches led researchers to build models of such online interactions using ontologies like FOAF, SIOC or SCOT.

This paper starts with a brief state of the art on these enhanced RDF-based representations. We will see that the graphs built using these ontologies have a great potential that is not fully exploited so far. Then, we present a new framework for applying SNA to RDF representations of social data. In particular, the use of graph models underlying RDF and SPARQL extensions enables us to extract efficiently and to parameterize the classic SNA features directly from these representations. We detail the computation of different measures of centrality [10] in RDF-based social network representations, using SPARQL extensions. An ontology called SemSNA is used to leverage social data and manage the life cycle of an analysis. Finally, we present some results based on real social data (we extracted thousands of FOAF profiles from users of flickr.com) and discuss the advantages, shortcomings and perspectives of such an approach to fully realize semantic social network analysis.

## 2. SNA and Semantic Web Technologies

Semantic web frameworks provide a graph model (RDF[1]), a query language (SPARQL[1]) and schema definition frameworks (RDFS[1] and OWL[1]) to represent and exchange knowledge online. These frameworks provide a whole new way of capturing social networks in much richer structures than raw graphs.

Recently, social interactions through web 2.0 social platforms have raised lots of attention in the semantic web community. Several ontologies are used to represent social networks [9]. FOAF[2] is used for describing people profiles, their relationships and their activities online. The properties of the RELATIONSHIP[3] ontology specialize the "knows" property of FOAF to type relationships in a social network more precisely (familial, friendship or professional relationships). The primitives of the SIOC[4] ontology extend FOAF in order to model more precisely online activities (e.g. posts in forums, blogs, etc). All these ontologies can be used and extended to link and reuse scenarios and data from web 2.0 community sites [4]. RDF based descriptions of social data provide a *rich typed graph* and offer a much more *powerful and significant way to represent online social networks than traditional models of SNA*.

Some researchers have applied classical SNA methods to the graph of acquaintance and interest networks respectively formed by the properties `foaf:knows` and `foaf:interest` [16]. In order to perform such an analysis, they chose to build *their own, untyped graphs* (each corresponding to one relationship "knows" or "interest") from the richer RDF descriptions of FOAF profiles. Too much knowledge is lost in this transformation and this knowledge could be used to parameterize social network indicators, improve their relevance and accuracy, filter their sources and customize their results.

In parallel, web 2.0 applications made social tagging popular: users tag resources of the web (e.g. media, blog posts etc.). Ontologies, like SCOT [12], have been designed to capture and exploit such activities and in parallel researchers have attempted to bridge folksonomies and ontologies to leverage the semantics of tags (see overview in [14]). *Once they are typed and structured, the relations between the tags and between the tags and the users also provide a new source of social networks.* A pioneering work by Peter Mika [15] proposed building such a network with a tripartite graph linking users, tags and resources with ternary edges. Mika chose not to work directly on this tripartite graph, and reduced this hypergraph into three bipartite graphs with untyped edges, two of them being the networks made of users who shared either the same tags or who tagged the same resources.

---

[1] Semantic Web, W3C, http://www.w3.org/2001/sw/

[2] http://www.foaf-project.org/

[3] http://vocab.org/relationship/

[4] http://sioc-project.org/

In all these experiments, researchers reduced the expressivity of the social network representations to simple untyped graphs, highlighting the lack of tools that can be applied directly on rich typed representation of social networks. We propose to exploit directly the RDF representations, in order to take advantage of the rich information they hold and in particular the typed relations that form these labelled graphs.

## 3. Semantic SNA Architecture

Among important results in SNA is the identification of sociometric features that characterize a network [9] (e.g. **community detection** shows the distribution of actors and activities, **centrality** highlights stategic positions). Figure 1 illustrates the abstraction stack we follow. We use the RDF graphs to represent social data, using existing ontologies together with specific domain ontologies if needed. Some social data are already readily available in a semantic format (RDF, RDFa, hCard µformat, etc.) and can be exploited straightforwardly. However, today, most of the data are still only accessible through APIs (e.g. flickr, Facebook, etc.) or by crawling web pages and need to be converted.

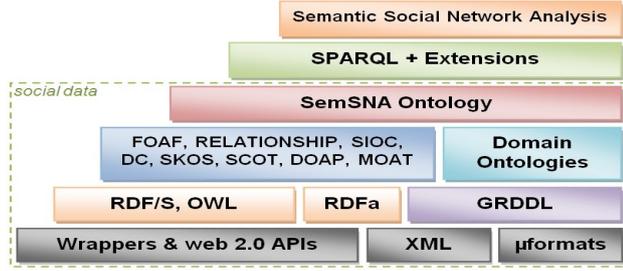

**Figure 1** Abstraction stack for semantic social network analysis

We designed SemSNA, an ontology that describes the SNA features (e.g. centrality). With this ontology, we can (1) abstract social network constructs from domain ontologies to apply our tools on existing schemas by having them extend our primitives; and we can (2) enrich the social data with new annotations such as the SNA indices that will be computed (e.g. centrality). These annotations enable us to manage more efficiently the life cycle of an analysis, by calculating only once the SNA indices and updating them incrementally when the network changes over time. SemSNA is detailed in section 3.2.

Based on this model, we propose SPARQL definitions to compute these classic SNA features and annotate the graph nodes consequently. The current test uses the semantic search engine CORESE [7] that supports powerful SPARQL extensions particularly well suited for the computation of the SNA features that require path computations [8]. These extensions will be detailed in the next section.

### 3.1 Paths in RDF graph with CORESE

The triples of an RDF description can be seen as the labelled arcs of an Entity-Relation graph [3]. Building on top of our work on graph-based representation and reasoning for RDF and OWL [8][3], we exploit the RDF graph representations of social networks.

**Definition of an ERGraph:** An ERGraph relative to a set of labels $L$ is a 4-tuple $G=(E_G, R_G, n_G, l_G)$ where :
- $E_G$ and $R_G$ are two disjoint finite sets respectively, of nodes called entities and of hyperarcs called relations.
- $n_G : R_G \rightarrow E_G^*$ associates to each relation a finite tuple of entities called the arguments of the relation. If $n_G(r)=(e_1,...,e_k)$ we note $n_G^i(r)=e_i$ the i$^{th}$ argument of $r$.
- $l_G : E_G \cup R_G \rightarrow L$ is a labelling function of entities and relations.

Intuitively, a mapping associates entities of a query ERGraph to entities of an ERGraph in a knowledge base of ERGraphs. Mapping entities of graphs is a fundamental operation for comparing and reasoning with graphs.

**Definition of an EMapping:** Let $G$ and $H$ be two ERGraphs, an EMapping from $H$ to $G$ is a partial function $M$ from $E_H$ to $E_G$ i.e. a binary relation that associates each element of $E_H$ with at most one element of $E_G$ ; not every element of $E_H$ has to be associated with an element of $E_G$ unless the mapping is total.

In addition to the mapping of entities linked by the graph, we may want the mapping to enforce that the relations of the graph being mapped are also preserved:

**Definition of an ERMapping:** Let $G$ and $H$ be two ERGraphs, an ERMapping from $H$ to $G$ is an EMapping $M$ from $H$ to $G$ such that: Let $H'$ be the SubERGraph of $H$ induced by $M^{-1}(E_G)$, $\forall r' \in R_{H'}$ $\exists r \in R_G$ such that $card(n_{H'}(r'))= card(n_G(r))$ and $\forall 1 \leq i \leq card(n_G(r))$, $M(n_{H'}^i(r'))= n_G^i(r)$. We call $r$ a support of $r'$ in $M$ and note $r \in M(r')$

Finally, the mapping of the labels of the entities or relations may be done according to an external schema of types of classes and properties, in other words, the mapping may take into account an ontology and in particular the pre-order relation defined by its taxonomical skeleton:

**Definition of an EMapping$_{<X>}$:** Let $G$ and $H$ be two ERGraphs, and $X$ be a binary relation over $L \times L$. An EMapping$_{<X>}$ from $H$ to $G$ is an EMapping $M$ from H to G such that $\forall e \in M^{-1}(E_G)$, $(l_G(M(e)), l_H(e)) \in X$.

By combining structural constraints and constraints on labelling, we now define the notion of ERMapping$_{<X>}$. In the special (but usual) case where $X$ is a preorder over $L$, the mapping defines the well-known notion of projection:

**Definition of an ERMapping$_{<X>}$:** Let $G$ and $H$ be two ERGraphs, and $X$ be a binary relation over $L \times L$. An ERMapping$_{<X>}$ $M$ from $H$ to $G$ is both an EMapping$_{<X>}$ from $H$ to $G$ and an ERMapping from $H$ to $G$ such that
- Let $H'$ be the SubERGraph of $H$ induced by $M^{-1}(E_G)$
- $\forall r' \in R_{H'}$ $\exists r \in M(r')$ such that $(l_G(r), l_H(r')) \in X$. we call $r$ a support$_{<X>}$ of $r'$ in $M$ and note $r \in M_{<X>}(r')$

**Definition of a Homomorphism$_{<X>}$:** Let $G$ and $H$ be two ERGraphs, a Homomorphism$_{<X>}$ from $H$ to $G$ is a total ERMapping$_{<X>}$ from $H$ to $G$ where $X$ is a preorder over $L$.

A Homomorphism$_{<X>}$ is also called a Projection. Mapping, especially total mapping, is a basic operation used in many more complex operations e.g. rule application, or query resolution. We use Homomorphism$_{<X>}$ implementations to operationnalize the SPARQL query language for RDF.

SPARQL extensions [1][2][8][13] have been implemented in the search engine CORESE [7][8]. These extensions enable us *to extract paths in RDF graphs* by specifying multiple criteria such as the type of the properties involved in the path with regular expressions, or edge directions or constraints on the vertices that paths go through.

The principle of the path extension consists of searching a property path of length more than one between two resources in the RDF graph. For this we extend the previous definitions to include paths in the graphs and mappings:

**Definition of an PERGraph:** An PERGraph relative to a set of labels $L$ is a 4-tuple $G=(E_G, R_G, P_G, n_G, l_G)$ where
- $G'=(E_G, R_G, n_G, l_G)$ is an ERGraph
- $P_G$ is a disjoint set from $E_G$ and $R_G$ of hyperarcs called paths.
- $n_G : R_G \cup P_G \rightarrow E_G^*$ associates to each relation and path a finite tuple of entities called the arguments of the relation or the path.
- $l_G : E_G \cup R_G \cup P_G \rightarrow L$ is a labelling function of entities, relations and paths.

**Definition of an PERMapping$_{<X>}$:** Let $G$ and $H$ be two PERGraphs, an PERMapping from $H$ to $G$ is an ERMapping$_{<X>}$ $M$ from $H$ to $G$ such that: Let $H'$ be the SubERGraph of $H$ induced by $M^{-1}(E_G)$, $\forall p' \in P_{H'} \exists p=(r_1,...,r_n) \in R_G^n$ such that
- $\forall 1 \leq i \leq card(n_G(r)), M(n_{H'}^i(r')) \in n_G(r_1) \cup n_G(r_n)$
- $\forall 1 \leq i \leq n \ (l_G(r_i), l_H(p')) \in X$.

To exploit this extension of the model, a syntactic convention enables us to specify that a path is searched, in our case we prefix the property variable with the character "$" instead of "?":

`<http://inria.fr/Fab>` **$path** `<http://orange.fr/Maya>`

In addition, following [13], it is possible to specify a regular expression on property names that the properties constituting a result path must match. For instance, to walk through relationships by means of a path of `foaf:knows` properties, we use the following regular expression from [8]:

`filter(match($path, star(foaf:knows)))`

As Corese supports RDFS entailment, it implements the *PERMapping$_{<X>}$* and thus takes into account sub-properties of the properties of the regular expression, unless specified otherwise.

An extension enables us to enumerate the triples belonging to the path that has been found. It uses the `graph` clause where the graph variable is a path variable. The example below enumerates the `foaf:knows` triples of the path and imposes that a least one resource knows `Michel`:

```
graph $path { ?x foaf:knows ?y }
?x foaf:knows <http://www.i3s.fr/Michel>
```

Paths in graph represent how nodes are interconnected and, in a social network, how resources are interacting. Betweenness centrality and community detection are among the more complex problem of SNA features computation and most of dedicated algorithms of graph theory are based on how actors are connected through *different path characteristics* (e. g. shortest paths) [9].

We will see how SPARQL can be used to query RDF social data with different perspectives. This querying, parameterized by typing, focuses automatically on specific path patterns, involving specific resource or property types. Such a tool enables us to detect rich relationships among the diversity of online interactions (e.g. two users annotated a same resource with a same concept). In next section 4 we will show that most of the SNA features can be directly computed using single SPARQL queries. In some cases, the implementation of complex SNA algorithms will require post-processing. Nevertheless, SPARQL is well suited for computing SNA features and helps extracting relevant parts of the graph and preparing required steps of algorithms.

Another advantage of querying social data with SPARQL is to ease the evolution of social platform models and the integration of new ones as they are expressed with ontologies. The web evolves very quickly, new social platforms with new features and usages frequently appear with new forms of social exchanges. As semantic web technologies are designed to tackle such a challenge, our approach is well adapted to the evolution of web platforms.

### 3.2 Measure Centrality with SPARQL

In this section, we present how we query the RDF social graph using path extensions in SPARQL to compute centralities and identify strategic positions using the three definitions of Freeman [10]. The path filtering features are used to parameterize the analysis: in order to deal with the diversity of social interactions that are captured online, we parameterize SNA features extraction with parameterized SPARQL queries and focus on relevant RDF sub-graph.

*3.2.1 Degree centrality*

The degree centrality considers people with higher degrees as more central, highlighting the local popularity of an actor in its neighborhood. The degree of a resource is the number of resources adjacent to it. The *n-degree* of a resource in a RDF graph is the number of paths of length *n* or less starting from or ending to this resource in other words, this is the number of sequences of *n* properties having this resource at an end.

We notate the parameterized degree: $deg_{<type,length>}(y)$ and extract it with the following query:

```
select ?y count(?x) as ?degree where{
{?x $path ?y
 filter(match($path, star(param[type])))
 filter(pathLength($path) <= param[length]) }
UNION
{?y $path ?x
 filter(match($path, star(param[type])))
 filter(pathLength($path) <= param[length]) }
} group by ?y
```

### 3.2.2 Closeness centrality

The closeness centrality of a resource represents its capacity to join (and to be reached by) any resource in a network. The closeness centrality of a node is the inverse sum of its shortest distances to each other resource [10]. with $n$ the number of nodes and $d(k,i)$ the length of a shortest path from $k$ to $i$ the closeness centrality is:

$$C_C(k) = \left[\sum_{i=1}^{n} d(k,i)\right]^{-1}$$

A shortest path in an RDF graph corresponds to a minimum number of triples that connect two resources.
We note $C_{c<type>}(y)$ the parameterized closeness centrality and we extract it with the following query:
```
select  ?y  ?to  pathLength($path)  as  ?length
sum(?length) as ?centrality where {
  ?y $path ?to
  filter(match($path, star(param[type]), 's'))
} group by ?y
```

### 3.2.3 Betweenness centrality

The betweenness centrality focuses on the capacity of a node to be an intermediary between any two other nodes. According to Freeman [10], the betweenness of $k$ for a couple of resource $(i, j)$ is the probability for $k$ to be on a shortest path from $i$ to $j$:

$$b_{ij}(k) = \frac{g_{ij}(k)}{g_{ij}}$$

with $g_{ij}$ the number of shortest paths between $i$ and $j$ and $g_{ij}(k)$ the number of shortest paths between $i$ and $j$ going through $k$. Then the betweenness centrality of $k$ in the whole network is the sum of its betweenness for all possible couples of resources:

$$C_B(k) = \sum_{i}^{n} \sum_{j=i}^{n} b_{ij}(k)$$

We compute $g_{count<type>}(from,to)$ the parameterized number of shortest paths between *from* and *to*, using the following query:
```
select  ?from  ?to  count($path)  as  ?nbPaths
where{
 ?from $path ?to
 filter(match($path, star(param[type]), 'sa'))
} group by ?from ?to
```
We compute $g_{count<type>}(b,from,to)$ the parameterized number of shortest paths between *from* and *to* going though *b* using the following query:
```
select  ?from  ?to  ?b count($path)  as  ?nbPaths
where{
 ?from $path ?to
 graph $path{?b param[type] ?j}
 filter(match($path, star(param[type]), 'sa'))
 filter(?from != ?b)
 optional { ?from param[type]::?p ?to }
 filter(!bound(?p))
} group by ?from ?to ?b
```
We consequently defined the parameterized betweenness:

$$B_{<type>}(b, from, to) = \frac{g_{count<type>}(b, from, to)}{g_{count<type>}(from, to)}$$

Finally, compute $C_{b<type>}(b)$ the parameterized betweenness centrality, by summing them:
Let $C_{b<type>}(b) \leftarrow 0$
```
For each pair <?from, ?to> connected by a
shortest path going through b :
```
$C_{b<type>}(b) += B_{<type>}(b,from,to)$

### 3.3 SemSNA: an ontology of SNA

SemSNA[5] is an ontology modelling concepts that are used in SNA such degree or centrality. It enables us to enrich social data by characterizing the structure of a social network, the strategic positions, and the way information flows.

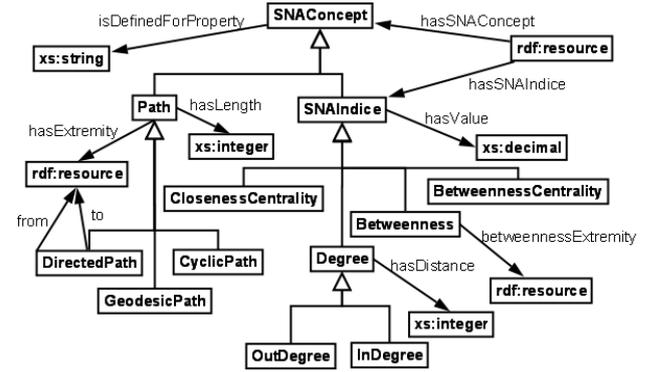

**Figure 2: Schema of SemSNA**

**Main primitives**: The main class is *SNAConcept*, used as super class for all SNA concepts. The property *isDefinedForProperty* indicates for which relationship (*i.e.* sub-network) an instance of SNA concept is defined. An SNA concept is attached to a social resource with the property *hasSNAConcept*. The class *SNAIndice* describes valued concepts such as centrality, and the associated value is set with the property *hasValue*. Again, an SNA index is attached to a resource with *hasSNAIndice* a sub property of has*SNAConcept*.

This basis enables us to describe the main indices of social network analysis. This first version of the ontology is focused on determining strategic positions that are represented by the three definitions of centrality proposed by Freeman [10].

**Centralities**: The degree centrality of a resource is the sum of its adjacent edges, *i.e.* its degree. We defined three subclasses of SNAIndice to describe a degree: *Degree*, *OutDegree* (is the sum of outgoing edges) and *InDegre*, (the *in degree* and *out degree* is sum of ingoing edges). An extension to the degree definition widens the neighbourhood to more distant connections of an actor indicated with *hasDistance* to describe the maximal distance that was considered. Then *BetweennessCentrality* and *ClosenessCentrality* classes are subclasses of *SNAIndice*, representing the eponymous centralities. The betweenness of a node for a given pair of nodes is modelled by the class *Betweenness*

---
[5] http://ns.inria.fr/semsna/2008/11/02/voc#

**Path definitions**: In order to annotate strategic paths we modelled different kind of paths using the subclass *Path* of *SNAConcept* and its three subclasses: *DirectedPath*, *GeodesicPath* and *CyclicPath*. The properties of the paths are *hasLength, pathExtremity* and *hasBetween,* used to precise respectively the length, the extremity and important intermediary nodes of a path. The sub-properties *from* and *to* of *pathExtremity* are used to describe the source and the destination of directed paths.

Leveraging the social graph with semantic SNA enables to query directly the social data in a cheaper way and to focus on important values of indices. Time-consuming queries can't be done in real time. We compute them as batch reporting and generate relevant SNA annotations enriching the graph in order to respond quickly to queries on demand.

## 4. Results

After having validated our approach on a toy example, we made some benchmarks on a larger social network. We extracted 2020 FOAF profiles from flickr.com, with 326617 `foaf:knows` properties linking 126308 `foaf:Person`, and tried our algorithms to identify most important betweenness centralities. We stopped after 50,000 graph pattern matchings. This process took less than 15 seconds on an Intel(R) Core(TM) Duo CPU T7300 @ 2.00 GHz with 2GB of RAM. Figure 3 shows the power law distribution of betweenness centralities higher than zero on this network, with the long tail capturing the important members. In the long tail we identified 7 actors playing a key role of intermediary and broker in the information flow. We also observed that 4 out of the 7 actors with the highest betweenness centralities for 50,000 matchings are among the five most important betweenness centralities for 10,000 matchings. Moreover, most of the lowest scores are not significant in both cases. These results suggest that *maybe* the exact algorithm for computing betweenness centralities could be turned into a good approximating algorithm by extrapolating from a subset of a social network, as stated by researchers [5].

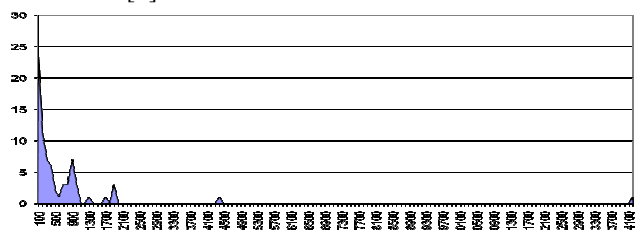

**Figure 3: Long tail distribution of the betweenness centralities**

## 5. Conclusion and Perspectives

Leveraging the semantics of social data in a machine readable format opens new perspectives for SNA and the enhancement of online social experiences. We proposed a semantic SNA stack to better exploit the rich representation of online social networks and leverage social data with SNA features.

Our perspectives include the adaptation of other algorithms in particular for community detection, and new semantic algorithms based on adaptation of classical SNA definitions. We will consequently extend SemSNA to semantically describe these new features and also help their querying extraction with inference rules on top of social data. Future work includes the development of iterative algorithms and methods to manage evolutions of ever-changing networks.

**Acknowledgements.** We thank the ANR for funding the ISICIL project ANR-08-CORD-011 that led to these results.